\documentclass[10pt]{article}
\usepackage{graphicx}
\usepackage{times}
\usepackage{amsmath}
\usepackage{amssymb}
\usepackage{units}
\usepackage{url}
\usepackage[noadjust,compress]{cite}
\usepackage{algorithm}
\usepackage{algorithmic}
\usepackage{nicefrac}
\usepackage{authblk}

\usepackage[left=2cm,right=2cm,%
           top=1.5cm,bottom=2cm,%
           includefoot,includehead]{geometry}

\newcommand{\MCW}{\ensuremath{\mathrm{MC}_\mathrm{W}}}

\newcommand{\MCMI}{\ensuremath{\mathrm{MC}_\mathrm{MI}}}

\newcommand{\discretised}[1]{#1^*}

\title{
  Evaluating Morphological Computation in Muscle and DC-motor Driven Models of Human Hopping
}
\author[1]{Keyan Ghazi-Zahedi}
\author[2]{Daniel F.\,B.\ Haeufle}
\author[1]{Guido Mont{\'u}far}
\author[2]{Syn Schmitt}
\author[1,3,4]{Nihat Ay}

\affil[1]{Max Planck Institute for Mathematics in the Science, Leipzig, Germany
    {\tt\small \{zahedi,montufar,nay\}@mis.mpg.de}}
\affil[2]{Stuttgart Research Centre for Simulation Technology and the Department
  of Sport and Movement Science, University of Stuttgart, Germany {\tt\small
    \{haeufle,schmitt\}@inspo.uni-stuttgart.de}}
\affil[3]{University of Leipzig, Leipzig, Germany}
\affil[4]{Santa Fe Institute, NM, USA}

\begin{document}
\twocolumn
\maketitle
\begin{abstract}
  In the context of embodied artificial intelligence, morphological computation
  refers to processes which are conducted by the body (and environment) that
  otherwise would have to be performed by the brain. Exploiting environmental
  and morphological properties is an important feature of embodied systems. The
  main reason is that it allows to significantly reduce the controller
  complexity. An important aspect of morphological computation is that it cannot
  be assigned to an embodied system per se, but that it is, as we show,
  behavior- and state-dependent. In this work, we evaluate two different
  measures of morphological computation that can be applied in robotic systems
  and in computer simulations of biological movement. As an example, these
  measures were evaluated on muscle and DC-motor driven hopping models. We show
  that a state-dependent analysis of the hopping behaviors provides additional
  insights that cannot be gained from the averaged measures alone. This work
  includes algorithms and computer code for the measures.

\end{abstract}

\section{Introduction}

Morphological computation (MC), in the context of embodied (artificial)
intelligence, refers to processes which are conducted by the body (and
environment) that otherwise would have to be performed by the brain
\cite{Pfeifer2006aHow-the-Body}. A nice example of MC is given by
Wootton~\cite{Wootton1992Functional-Morphology-of-Insect} (see p.~188), who
describes how ``active muscular forces cannot entirely control the wing shape in
flight. They can only interact dynamically with the aerodynamic and inertial
forces that the wings experience and with the wing's own elasticity; the
instantaneous results of these interactions are \emph{essentially} determined by
the architecture of the wing itself [\ldots]''

MC is relevant in the study of biological and robotic systems. In robotics, a
quantification of MC can be used e.g.~as part of a reward function in a
reinforcement learning setting to encourage the outsourcing of computation to
the morphology, thereby enabling complex behaviors that result from comparably
simple controllers. The relationship of embodiment and controller complexity has
been recently studied in~\cite{Montufar2015aA-Theory}. MC measures can also be
used to evaluate the robot's morphology during the design process. For
biological systems, energy efficiency is important and an evolutionary
advantage. Exploiting the embodiment can lead to more energy efficient
behaviors, and hence, MC may be a driving force in evolution.

In biological systems, movements are typically generated by muscles.
Several simulation studies have shown that the muscles' typical non-linear
contraction dynamics can be exploited in movement generation with very simple
control strategies~\cite{Schmitt2015a-SoftRobotics}. Muscles improve movement
stability in comparison to torque driven models~\cite{Soest1993a-Contribution}
or simplified linearized muscle models (for an overview
see~\cite{Haeufle2012aIntegration}). Muscles also reduce the influence of the
controller on the actual kinematics (they can act as a low-pass filter). This
means that the hopping kinematics of the system is more pre-determined with
non-linear muscle characteristics than with simplified linear muscle
characteristics~\cite{Haeufle2012aIntegration}. And finally, in hopping
movements, muscles reduce the control effort (amount of information required to
control the movement) by a factor of approximately 20 in comparison to a
DC-motor driven movement~\cite{Haeufle2014aQuantifying}.

In view of these results we expect that MC plays an important role in the
control of muscle driven movement. To study this quantitatively, a suitable
measure for MC is required. There are several approaches to formalize
MC~\cite{Hauser2011aTowards,Ruckert2013aStochastic,Polani2011aAn-informational}.
In our previous work we have focused on an agent-centric perspective of
measuring MC~\cite{Zahedi2013aQuantifying} and we have applied an information
decomposition of the sensorimotor loop to measure and better understand
MC~\cite{Ghazi-Zahedi2015bQuantifying}. Both publications used a binary toy
world model to evaluate the measures. With this toy model, it was possible to
show that these measures capture the conceptual idea of MC and, in consequence,
that they are candidates to measure MC in more complex and more realistic
systems.

The goal of this publication is to evaluate two measures of MC on biologically
realistic hopping models. With this, we want to demonstrate their applicability
in non-trivial, realistic scenarios. Based on our previous findings (see above),
we hypothesize that MC is higher in hopping movements driven by a non-linear
muscle compared to those driven by a simplified linear muscle or a DC-motor, for
the following reason. Our experiments show that a state-dependent analysis of MC
for the different models leads to insights, which cannot be gained from the
averaged measures alone.

Furthermore, we provide detailed instructions, including
MATLAB\textsuperscript{\textregistered} code, on how to apply these measures to
robotic systems and to computer simulations. With this, we hope to provide a
tool for the evaluation of MC in a large variety of
applications.


The quantifications of MC require a formal representation of the sensorimotor
loop (see Fig.~\ref{fig:sml}), which is introduced in the next section as far as
it is required to understand the remainder of this work. For further
information, the reader is referred to~\cite{Klyubin2004aOrganization,Zahedi2010aHigher,Zahedi2013aQuantifying,Ay2013aAn-Information}.

\section{The Sensorimotor Loop}
\label{sec:sml}
The conceptual idea of the sensorimotor loop is similar to the basic control
loop systematics, which is the basis of robotics and also of computer
simulations of human movement. In our understanding, a cognitive system consists
of a brain or controller, which sends signals to the system's actuators, thereby
affecting the system's environment. We prefer to capture the body and
environment in a single random variable named \emph{world}. This is consistent
with other concepts of agent-environment distinctions. An example for such a 
distinction can be found in the context of reinforcement learning,
where the environment (world) is everything that cannot be changed arbitrarily
by the agent \cite{Sutton1998aReinforcement}. A more thorough
discussion of the brain-body-environment distinction can be found in
\cite{Uexkuell1957aA-Stroll,Ay2015aThe-Umwelt} and a more recent discussion can
be found in \cite{Clark1996aBeing}. A brief example of a world, based on a robot
simulation, is given below. The loop is closed as the system acquires
information about its internal state (e.g.~current pose) and its world through
its sensors.

\begin{figure}[t]
  \begin{center}
    \includegraphics[width=0.45\textwidth]{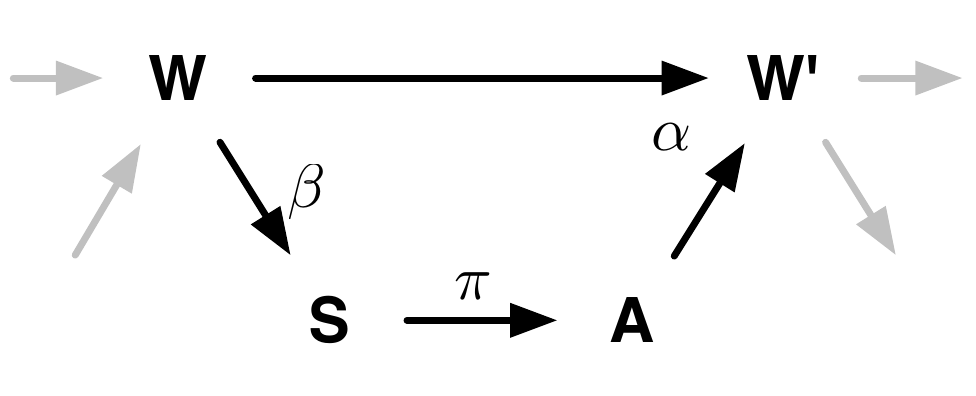}
  \end{center}
  \caption{Causal model of the sensorimotor loop. This figure depicts a single
    step of an embodied, reactive system's sensorimotor loop. A detailed
    explanation is given in Sec.~\protect\ref{sec:sml}}\label{fig:sml}
\end{figure}

For simplicity, we only discuss the sensorimotor loop for reactive systems. This
is plausible, because behaviors which exploit the embodiment (e.g.~walking,
swimming, flying) are typically reactive. This leaves us with three (stochastic)
processes $S(t)$, $A(t)$, and $W(t)$, $t\in\mathbb{N},$ that constitute the
sensorimotor loop (see Fig.~\ref{fig:sml}), which take values $s$, $a$, and $w$,
in the sensor, actuator, and world state spaces (their respective domains will
be clear from the context). The directed edges (see Fig.~\ref{fig:sml}) reflect
causal dependencies between these random variables. We consider time to be
discrete, i.e., $t\in\mathbb{N}$ and are interested in what happens in a single
time step. Therefore, we use the following notation. Random variables without
any time index refer to some fixed time $t$ and primed variables to time $t+1$,
i.e., the two variables $S$, $S'$ refer to $S(t)$ and $S(t+1)$.


Starting with the initial distribution over world states, denoted by $p(w)$, the
sensorimotor loop for reactive systems is given by three conditional probability
distributions, $\beta$, $\alpha$, $\pi$, also referred to as kernels.
The sensor kernel, which determines how the agent perceives the world, is
denoted by $\beta(s|w)$, the agent's controller or policy is denoted by
$\pi(a|s)$, and finally, the world dynamics kernel is denoted by
$\alpha(w'|w,a)$.

To understand the function of the world dynamics kernel $\alpha(w'|w,a)$ it is
useful to think of a robotic simulation. In this scenario, the world state $W$
is the state of the simulator at a given time step, which includes the pose of all objects,
their velocities, applied forces, etc. The actuator state $A$ is the value that the controller passes on to the physics engine prior to the next
physics update. Hence, the world dynamics kernel $\alpha(w'|w,a)$ is closely
related to the forward model that is known in the context of robotics and
biomechanics.

Based on this notation, we can now formulate quantifications of MC in the next
section.

\section{Quantifying Morphological Computation}
In the introduction, we stated that MC relates to the computation that the body
(and environment) performs that otherwise would have to be conducted by the
controller (or brain). This means that we want to measure the extent to which
the system's behavior is the result of the world dynamics (i.e., the body's
internal dynamics and it's interaction with its world) and how much of the
behavior is determined by the policy $\pi$ (see Fig.~\ref{fig:sml}).

In our previous publication \cite{Zahedi2013aQuantifying} we have defined two
concepts to quantify MC, from which the two measures below are taken and derived.

\subsection{Morphological computation as conditional mutual information ($\mathrm{MC}_\mathrm{W}$)}
The first quantification, that is used in this work, was introduced in
\cite{Zahedi2013aQuantifying}. The
idea behind it can be summarized in the following way. The world dynamics kernel
$\alpha(w'|w,a)$ captures the influence of the actuator signal $A$ and the
previous world state $W$ on the next world state $W'$. A complete lack of
MC would mean that the behavior of the system is entirely
determined by the system's controller, and hence, by the actuator state $A$. In
this case, the world dynamics kernel reduces to $p(w'|a)$.
Every
difference from this assumption means that the previous world state $W$ had an
influence, and hence, information about $W$ changes the distribution over the
next world states $W'$. The discrepancy of these two distributions can be measured
with the average of the Kullback-Leibler divergence
$D_\mathrm{KL}(\alpha(w'|w,a)||p(w'|a))$, which is also known as the
conditional mutual information $I(W';W|A)$. This distance 
is formally given by (see also Alg.~\ref{alg:mc_w} in
App.~\ref{app:algorithms})
$$
\mathrm{MC}_\mathrm{W} := \sum_{w',w,a} p(w',w,a)
\log_2\frac{\alpha(w'|w,a)}{p(w'|a)}. \eqno{(1)}
$$

\subsection{Morphological computation as comparison of behavior and controller
complexity ($\mathrm{MC}_\mathrm{MI}$)}
The second quantification follows concept one of
\cite{Zahedi2013aQuantifying}. The assumption that underlies this concept is
that, for a given behavior,
MC decreases with an increasing effect of the action
$A$ on the next world state $W'$. The corresponding measure
$\mathrm{MC}_{\mathrm{A}} \propto -I(W';A|W)$ cannot be used in systems with
deterministic policy, because for these systems $I(W';A|W) = 0$ (see
App.~\ref{app:zero}).
Therefore, for this publication, we require
an adaptation that operates on world states and is applicable
to deterministic systems.

The new measure compares the complexity of the behavior with the
complexity of the controller. The complexity of the behavior can be measured by
the mutual information of consecutive world states, $I(W';W)$, and the
complexity of the controller can be measured by the mutual information of sensor
and actuator states, $I(A;S)$, for the following reason. The mutual information
of two random variables can also be written as difference of entropies:
\begin{align*}
I(X;Y) & = H(X) - H(X|Y)\\
H(X)   & = - \sum_x p(x) \log_2 p(x)\\
H(X|Y) & = - \sum_{x,y} p(x,y) \log_2 p(x|y),
\end{align*}
which, applied to our setting, means that the mutual
information $I(W';W)$ is high, if we have a high entropy over world states $W'$
(first term) that are highly predicable (second term). Summarized, this means
that the mutual information $I(W';W)$ is high if the system shows a diverse but
non-random behavior. Obviously, this is what we would like to see in an embodied
system. On the other hand, a system with high MC should
produce a complex behavior based on a controller with low complexity. Hence, we
want to reduce the mutual information $I(A;S)$, because this either means that
the policy has a low diversity in its output (low entropy over actuator states
$H(A)$) or that there is only a very low correlation between sensor states $S$
and actuator states $A$ (high conditional entropy $H(A|S)$). Therefore, we
define the second measure as the difference of these two terms, which is (see also Alg.~\ref{alg:mc_mi} in Sec.~\ref{app:algorithms})
$$
\mathrm{MC}_\mathrm{MI} = I(W';W) - I(A;S) .\eqno{(2)}
$$
For deterministic systems, as those studied in this work, the two measures are
closely related. In particular, it holds that $\mathrm{MC}_\mathrm{W} -
\mathrm{MC}_\mathrm{MI}  = H(A|W')$ (see App.~\ref{app:relation}). The
inequality $\mathrm{MC}_\mathrm{W} \geq \mathrm{MC}_\mathrm{MI}$ may not be
satisfied always, because discretization can introduce stochasticity.

Note that in the case of a passive observer, i.e., a system that observes the
world but in which there is no causal dependency between the action and the next
world state (i.e., missing connection between $A$ and $W'$ in
Fig.~\ref{fig:sml}), the controller complexity $I(A;S)$ in Eq.~(2) will reduce
the amount of MC measured by \MCMI, although the actuator state does not
influence the world dynamics. This might be perceived as a potential
shortcoming. In the context discussed in this paper, e.g.~data recorded from
biological or robotic systems, we think that this will not be an issue.

The next section introduces the hopping models on which the two measures are
evaluated.



\section{Hopping models}
In a reduced model, hopping motions can be described by a one-dimensional
differential equation~\cite{Haeufle2010aThe-role}: 
$$
m\ddot{y}=-mg+\begin{cases}
0 & y>l_{0}\text{\,\,\,\,\,\ flight phase}\\
F_{\text{L}} & y\leq l_{0}\text{\,\,\,\,\,\ ground contact}
\end{cases}\,\,\,,\label{eq:HuepfenDGLmath}
\eqno{(3)}
$$
where the point mass $m=\unit[80]{kg}$ represents the total mass
of the hopper which is accelerated by the gravitational force ($g=-\unit[9.81]{m/s^{2}}$)
in negative $y$-direction. An opposing leg force $F_{\text{L}}$
in positive $y$-direction can act only during ground contact ($y\leq l_{0}=\unit[1]{m}$).
Hopping motions are then characterized by alternating flight and stance
phases. For this manuscript, we investigated three different models
for the leg force. All models have in common, that the leg force depends
on a control signal $u(t)$ and the system state $y(t)$, $\dot{y}(t)$: 
$F_{\text{L}}=F_{\text{L}}(u(t),y(t),\dot{y}(t))$,
meaning, that the force modulation partially depends on the controller
output $u(t)$ and partially on the dynamic characteristics, or material
properties of the actuator. The control parameters of all three models
were adjusted to generate the same periodic hopping height of $\max(y(t))=\unit[1.070]{m}$.
All models were implemented in MATLAB\textsuperscript{\textregistered}
Simulink\textsuperscript{\texttrademark} (Ver2014b) and solved with
ode45 Dormand-Prince solver with absolute and relative tolerances
of $10^{-12}$. The models were solved and integrated in time for $T=\unit[8]{s}$ and
model output was generated at $\unit[1]{kHz}$ sampling frequency.

\begin{figure}
  \begin{centering}
    \includegraphics[width=\columnwidth]{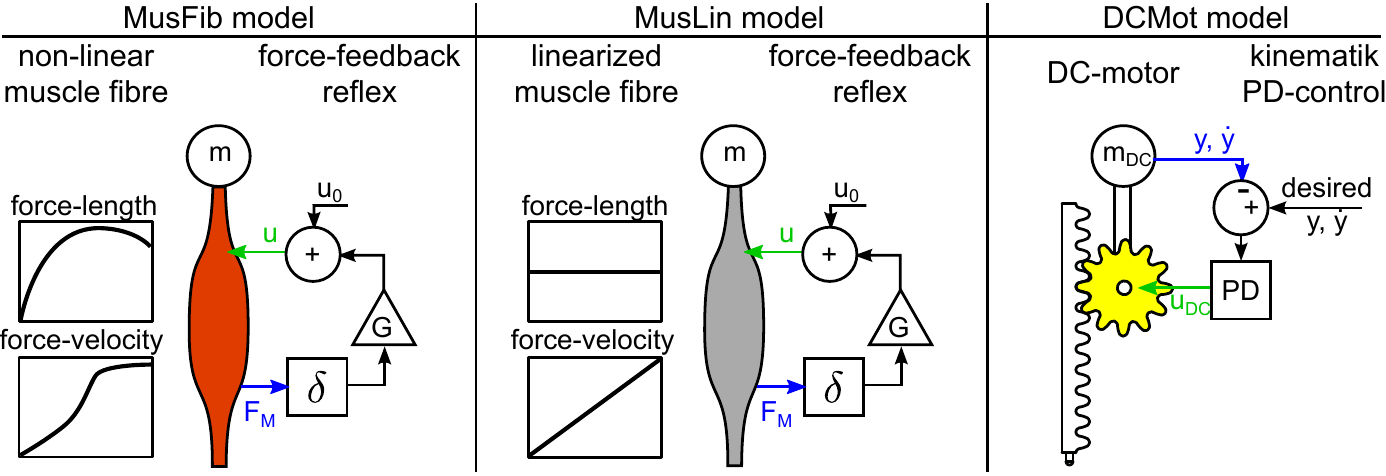}
    \par\end{centering}
  \protect\caption{Hopping models. The MusFib model considers the non-linear
    contraction dynamics of active muscle fibers and is driven by a mono-synaptic
    force-feedback reflex. The MusLin model only differs in the contraction
    dynamics, where the force-length relation is neglected and the force-velocity
    relation is approximated linearly. The DCMot model generates the leg force with
    a DC-motor. It is controlled by a proportional-differential controller (PD),
    enforcing the desired trajectory. The desired trajectory is the recorded
    trajectory from the MusFib Model. The sensor signals are shown as blue arrows,
    the actuator control signals are shown as green arrows. In case of the muscle
    models, the sensor signal is the muscle force $F_{M}$ and the actuator control
    signal is the neural muscle stimulation $u$. In case of the DC-motor model, the
    sensor signals are the position and velocity of the mass, and the actuator
    control signal is the motor armature voltage $u_{DC}$.}
\end{figure}

\subsection{Muscle-Fiber model (MusFib)}
A biological muscle generates its active force in muscle fibers whose
contraction dynamics are well studied. It was found that the contraction
dynamics are qualitatively and quantitatively (with some normalizations) very
similar across muscles of all sizes and across many species. In the MusFib model,
the leg force is modeled to incorporate the active muscle fibers' contraction
dynamics. The model has been motivated and described in detail elsewhere
\cite{Haeufle2010aThe-role,Haeufle2012aIntegration,Haeufle2014aQuantifying}. In
a nutshell, the material properties of the muscle fibers are characterized by
two terms modulating the leg force 
$$
F_{L,\text{MusFib}}=a(t)F_{\text{fib}}(l_{M},\dot{l}_{M}).\label{eq:HillModell}
\eqno{(4)}
$$
The first term $a(t)$ represents the muscle activity. The activity
depends on the neural stimulation of the muscle $0.001\le u(t)\le1$
and is governed by biochemical processes modeled as a first-order
ODE called activation dynamics 
$$
\dot{a}=\frac{1}{\tau}\left(u-a\right),\label{eq:ForceFeedback-1}
\eqno{(5})
$$
with the time constant $\tau=\unit[10]{ms}$. The second term in Eq.~(4)
$F_{\text{fib}}$
considers the force-length and force-velocity relation of biological
muscle fibers. It is a function of the system state, i.e., the muscle
length $l_{M}=y$ and muscle contraction velocity $\dot{l}_{M}=\dot{y}$
during ground contact $y\le l_{0}$ and constant $l_{M}=l_{0}$ $\dot{l}_{M}=0$
during flight $y > l_{0}$:

$$
F_{\text{fib}}=
\begin{aligned}[t]
& F_{\max}\cdot\exp\left(-c\left\lvert
    \frac{l_{M}-l_{\text{opt}}}{l_{\text{opt}}w}\right\lvert ^{3}\right)\\
& \times\begin{cases}
\frac{\dot{l}_{M,\text{max}}+\dot{l}_{M}}{\dot{l}_{M,\text{max}}-K\dot{l}_{M}} & \dot{l}_{M}>0\\
N+(N-1)\frac{\dot{l}_{M,\text{max}}-\dot{l}_{M}}{-7.56K\dot{l}_{M}-\dot{l}_{M,\text{max}}} & \dot{l}_{M}\leq0
\end{cases}\,\,\,.%
\end{aligned}\eqno{(6)}
$$
Here we use a maximum isometric muscle force $F_{\text{max}}=\unit[2.5]{kN}$,
an optimal muscle length $l_{\text{opt}}=\unit[0.9]{m}$, force-length
parameters $w=\unit[0.45]{m}$ and $c=30$, and force-velocity parameters
$\dot{l}_{\max}=\unit[-3.5]{ms^{-1}}$, $K=1.5$, and
$N=1.5$~\cite{Haeufle2010aThe-role}.

In this model, periodic hopping is generated with a controller representing
a mono-synaptic force-feedback. The neural muscle stimulation 
$$
u(t)=G\cdot F_{L,\text{MusFib}}(t-\delta)+u_{0}\eqno{(7)}
$$
is based on the time delayed ($\delta=\unit[15]{ms}$) muscle fiber
force $F_{L,\text{MusFib}}$. The feedback gain is $G=2.4/F_{\max}$
and the stimulation at touch down $u_{0}=0.027$. 

This model neither considers leg geometry nor tendon elasticity and
is therefore the simplest hopping model with muscle-fiber-like contraction
dynamics. The model output was the world state
$w(t)=(y(t),\dot{y}(t),\ddot{y}(t))$, the sensor state
$s(t)=F_{L\text{,MusFib}}(t)$, and the actuator control command $a(t)=u(t)$.
For this model, these are the values that the random variables $W$, $S$, and $A$ take at each time step.

\subsection{Linearized Muscle-Fiber model (MusLin)}
This model differs from the model MusFib only in the representation of the
force-length-velocity relation, i.e.,
$F_{L,\text{MusLin}}=a(t)F_{\text{lin}}(\dot{l}_{M})$ (see Eq.~(6)). More
precisely, the force-length relation is neglected and the force-velocity
relation is approximated linearly
$$
F_{\text{lin}}=1\cdot(1-\mu\dot{l}_{M}), \eqno{(8)}
$$
with $\mu=\unit[0.25]{m/s}$. Feedback gain $G=0.8/F_{\max}$ and stimulation
at touch down $u_{0}=0.19$ were chosen to achieve the same hopping height as the
MusFib model.

\subsection{DC-Motor model (DCMot)}
An approach to mimic biological movement in a technical system (robot)
is to track recorded kinematic trajectories with electric motors and
a PD-control approach. The DCMot model implements this approach (slightly
modified from \cite{Haeufle2014aQuantifying}).
The leg force generated by the DC-motor was modeled as
$$
F_{L,\text{DCMot}}=\gamma T_{DC}=\gamma k_{T}I_{DC}, 
\eqno{(9)}
$$
where $k_{T}=\unit[0.126]{Nm/A}$ is the motor constant, $I_{DC}$
the current through the motor windings, $\gamma=100:1$ the ratio
of an ideal gear translating the rotational torque $T_{DC}$ and movement
$\dot{\varphi}(t)=\gamma\dot{y}(t)$ of the motor to the translational
leg force and movement required for hopping. The electrical characteristics
of the motor can be modeled as

$$
\dot{I}_{DC} = \frac{1}{L}\left(u_{DC}-k_{T}\gamma\dot{y}(t)-RI_{DC}\right),
\eqno{(10)}
$$
where $-\unit[48]{V}\le u_{DC}\le\unit[48]{V}$ is the armature voltage
(control signal), $R=\unit[7.19]{\Omega}$ the resistance, and $L=\unit[1.6]{mH}$
the inductance of the motor windings. The motor parameters were taken
from a commercially available DC-motor commonly used in robotics applications
(Maxon EC-max 40, nominal Torque $T_{\text{nominal}}=\unit[0.212]{Nm}$).
As this relatively small motor would not be able to lift the same
mass, the body mass was adapted to guarantee comparable accelerations
$$
m_{DC}=\frac{\gamma
  T_{\text{nominal}}}{F_{\max}}m=\unit[0.68]{kg}.\label{eq:m_DC}
\eqno{(11)}
$$

The recorded kinematic trajectory $y_{\text{rec}}(t)$ and
$\dot{y}_{\text{rec}}(t)$ during ground contact was taken from the periodic
hopping trajectory of the MusFib model. This trajectory was enforced with a
PD-controller $$ u_{DC}(t) =
K_{P}(y_{\text{rec}}(t)-y(t))+K_{D}(\dot{y}_{\text{rec}}(t)-\dot{y}(t))\label{eq:PDcontroller}
\eqno{(12}) $$ with feedback gains $K_{P}=\unit[5000]{V/m}$ and
$K_{D}=\unit[500]{Vs/m}$. 

This model is the simplest implementation of negative feedback control
that allows to enforce a desired hopping trajectory on a technical
system. The model output was the world state $w(t)=(y(t),\dot{y}(t),\ddot{y}(t))$,
the sensor state $s(t)=(y(t),\dot{y}(t))$, and the actuator
control command $a(t)=u_{DC}(t)$. 

\section{Experiments}
This section discusses the experiments that were conducted with the hopping
models and the preprocessing of the data. Algorithms for the calculations are
provided in the appendix (Sec.~\ref{app:algorithms}) and implemented
MATLAB\textsuperscript{\textregistered} code can be downloaded from
\url{http://github.com/kzahedi/MC/} (commit c332c18, 30.~Nov.~2015). A
\verb!C++! implementation is available at
\url{http://github.com/kzahedi/entropy/}.

At this stage, the measures operate on discrete state spaces (see Eqs.~(1)--(3)
and Algs.~\ref{alg:mc_w}-\ref{alg:mc_mi}). Hence, the data was discretised in
the following way. To ensure the comparability of the results, the domain (range
of values) for each variable (e.g.~the position $y$) was calculated over all
hopping models.
Then, the data of each variable was discretised into 300 values (bins). The
algorithm for the discretisation is described in Alg. 1. Different binning
resolutions were evaluated and the most stable results were found for more than
100 bins. Finding the optimal binning resolution is a problem of itself and
beyond the scope of this work. In practice, however, a reasonable binning can be
found by increasing the binning until further increase has little influence on
the outcome of the measures.

The possible range of actuator values are different for the motor and muscle
models. For the muscle models, the values are in the unit interval, i.e.,
$a(t)\in[0,1]$, whereas the values for the motor can have higher values (see
above). Hence, to ensure comparability, we normalized the actions of the motor
to the unit interval before they were discretized.

The hopping models are deterministic, which means that only a few hopping cycles
are necessary to estimate the required probability distributions.
To ensure comparability of the results, 
we parameterized the hopping models to achieve the same hopping height.


\section{RESULTS}
Tab.~\ref{tab:results_on_data} shows the value of the two MC
measures for the three hopping models.
\begin{table}[h!]
  \begin{center}
    \begin{tabular}{l|c|c|c}
           & MusFib & MusLin & DCMot \\
      \hline
      $\mathrm{MC}_\mathrm{W}$  & 7.219 bits & 4.975 bits & 4.960 bits \\
      $\mathrm{MC}_\mathrm{MI}$ & 7.310 bits & 5.153 bits & 4.990 bits
    \end{tabular}
  \end{center}
  \caption{Numerical results on the hopping models for \MCW{} (see Eq.~(1)) and
    \MCMI{} (see Eq.~(2)).
  }
  \label{tab:results_on_data}
\end{table}
Compared to the MusFib model, the two other models result in significantly
lower measurement of MC ($\approx30\%$ less).
This result complements previous findings showing that the
minimum information required to generate hopping is reduced by the material
properties of the non-linear muscle fibers compared to the DC-motor driven
model~\cite{Haeufle2014aQuantifying}.

\begin{figure}[h]
  \begin{center}
    \includegraphics[clip, trim=0.8cm .2cm 0cm 0cm,width=\columnwidth]{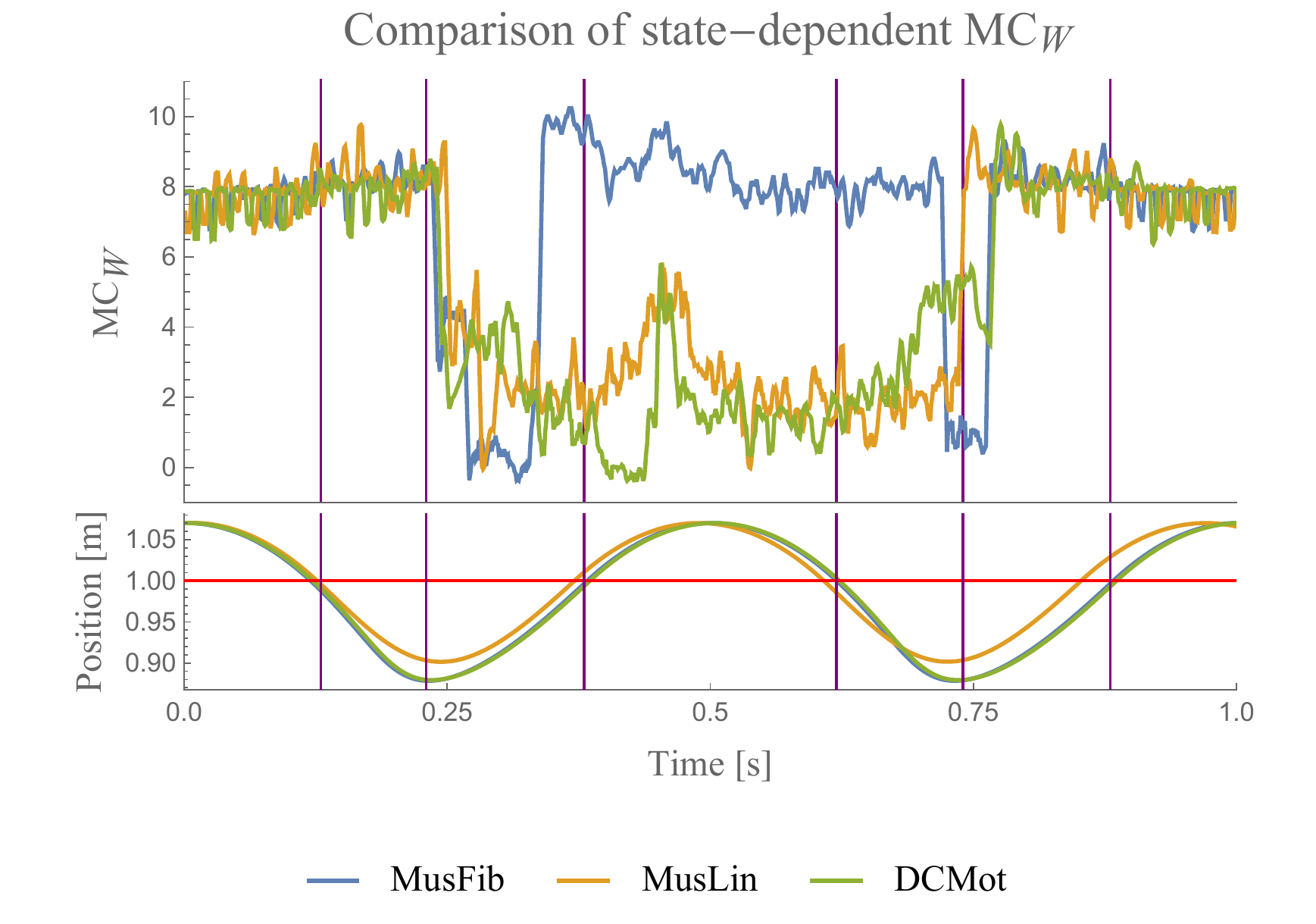}
  \end{center}
  \caption{Comparison of state-dependent MC for \MCW{} on the three hopping
    models. The lower plots in each Figure visualize the hopping position (out
    of proportion for better visibility). The red line indicates stance and
    flight phases. The full data is shown in
    Fig.~\protect\ref{fig:information_dynamics}. The plots only show a small
    fraction of the recorded data. For better readability, all the plots for MC
    are smoothened with a moving average of block size 5.
  }\label{fig:cmp}
\end{figure}

\begin{figure}[p]
  \begin{center}
    \includegraphics[width=0.9\textwidth]{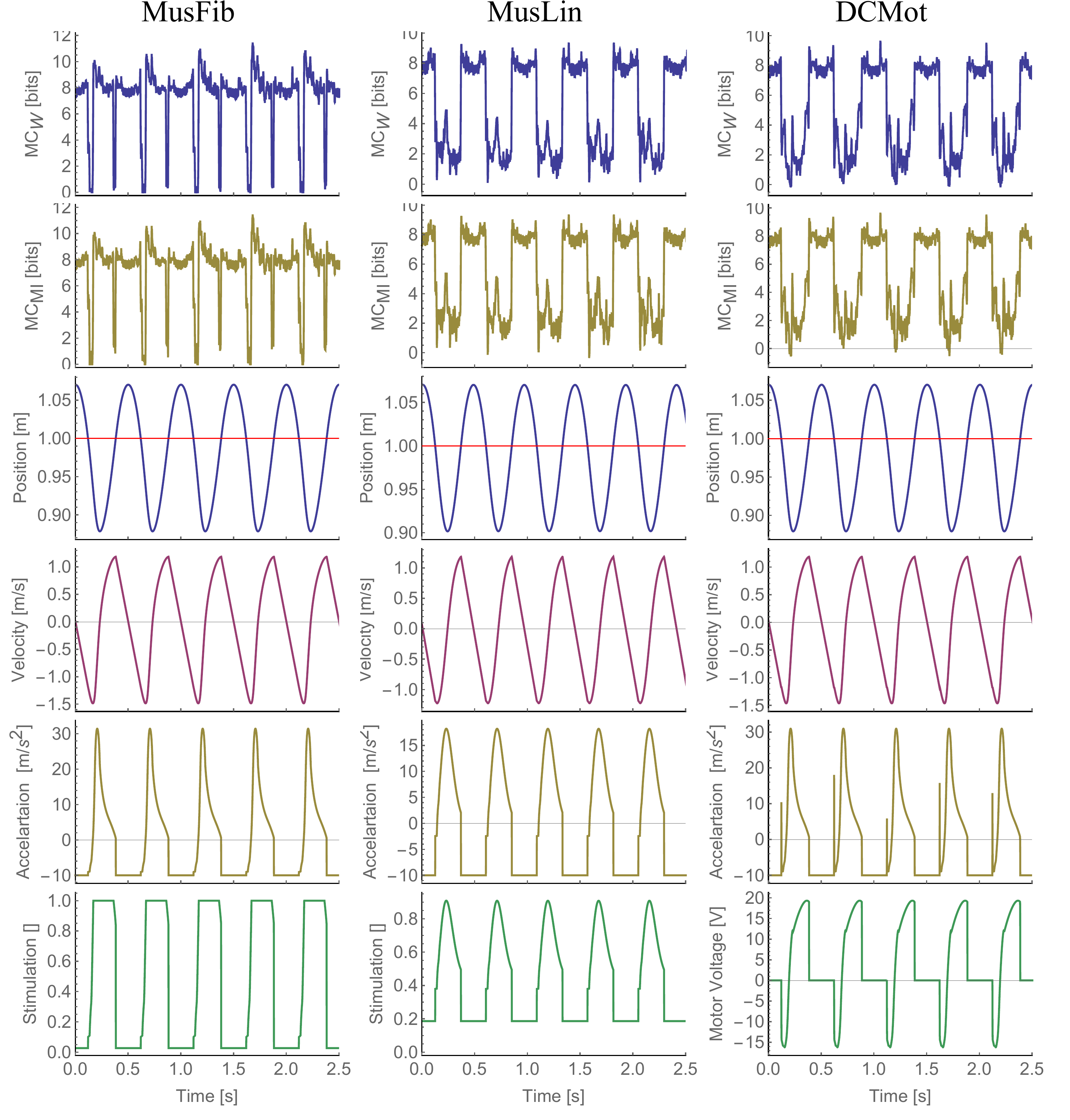}
  \end{center}
  \caption{Plots of state-dependent MC (first two rows) and the world state
    (following four rows) for all muscle models. The red line in the position
    plot indicates the time steps at which the hopper touches ground (position
    is below the red line).}\label{fig:information_dynamics}
\end{figure}

This also confirms previous findings that the non-linear contraction dynamics
reduces the influence of the controller on the actual hopping kinematics in
comparison to a linearized muscle model
\cite{Haeufle2010aThe-role,Haeufle2012aIntegration}. To better understand the
differences of the models, we plotted the state-dependent MC (see
Alg.~\ref{alg:sd_mc_w}). Fig.~\ref{fig:cmp} shows the values of
$\mathrm{MC}_{\mathrm{W}}$ for each state of the models during two hopping
cycles. We chose to discuss $\mathrm{MC}_{\mathrm{W}}$ only, because the plots
of $\mathrm{MC}_{\mathrm{W}}$ and \MCMI{} are very similar, and hence, a
discussion of the state-dependent \MCMI{} will not provide any additional
insights. The plots for all models and the entire data are shown in
Fig.~\ref{fig:information_dynamics}.

The orange line shows the state-dependent MC for the linear muscle model
(MusLin), and finally, the blue line shows the values for the non-linear muscle
model (MusFib). The green line shows the state-dependent MC for the motor model
(DCMot). In the figure, the lower lines show the position $y$ of the center of
mass over time. The DCMot model is parameterized to follow the trajectory of the
MusFib model (see Eq.~(12)), which is why the blue and green position plots
coincide. The original data is shown in Fig.~\ref{fig:information_dynamics}.
There are basically three phases, which need to be distinguished (indicated by
the vertical lines). First, the flight phase, during which the hopper does not
touch the ground (position plots are above the red line), second, the
deceleration phase, which occurs after landing (position is below the red line
but still declining), and finally, the acceleration phase, in which the position
is below the red line but increasing.

The first observation is that MC is equal for all models during most of the
flight phase (position above the red line) and that it seems to be proportional
to the velocity of the systems. During flight, the behavior of the system is
governed only by the interaction of the body (mass, velocity) and the
environment (gravity) and not by the actuator models. This explains why the values
coincide for the three models.

For all models, MC drops as soon as the systems touch the ground. DCMot and
MusLin reach their highest values only during the flight phase, which can be
expected at least from a motor model that is not designed to exploit MC. The
graphs also reveal that the MusLin model shows slightly higher MC around
mid-stance phase, compared to the DCMot model. For the non-linear muscle model,
the behavior is different. Shortly after touching the ground, the system shows a
strong decline of MC, which is followed by a strong incline during the
deceleration with the muscle. Contrary to the other two models, the non-linear
muscle model MusFib shows the highest values when the muscle is contracted the
most (until mid-stance). This is an interesting result, as it shows that the
non-linear muscle is capable of showing more MC while the muscle is operating,
compared to the flight phase, in which the behavior is only determined by the
interaction of the body and environment.

\section{CONCLUSIONS}
\label{sec:conclusions}
This work presented two different quantifications of MC including algorithms and
MATLAB\textsuperscript{\textregistered} code to use them. We demonstrated their
applicability in experiments with non-trivial, biologically realistic hopping
models and discussed the importance of a state-based analysis of morphological
computation. The first quantification, \MCW{}, measures MC as the conditional
mutual information of the world and actuator states. Morphological computation
is the additional information that the previous world state $W$ provides about
the next world state $W'$, given that the current actuator state $A$ is known.
The second quantification, \MCMI{}, compares the behavior and controller
complexity to determine the amount of MC.

The numerical results of the two quantifications confirm our hypothesis that the
MusFib model should show significantly higher MC, compared to the two other
models (MusLin, DCMot). We also showed that a state-dependent analysis of MC
leads to additional insights. Here we see that the non-linear muscle model is
capable of showing significantly more morphological computation in the stance
phase, compared to the flight phase, during which the behavior is only
determined by the interaction of the body and environment. This shows that
morphological computation is not only behavior-, but also state-dependent.
Future work will include the analysis of additional behaviors, such as walking
and running, for which we expect, based on the findings of this work, to see a
more morphological computation of the non-linear muscle model MusFib.



\section*{APPENDIX}
\section{Algorithms}
\label{app:algorithms}
This section presents the algorithms in pseudo-code. The
MATLAB\textsuperscript{\textregistered} code that was used for this publication
can be downloaded from \url{http://github.com/kzahedi/MC/} (commit c332c18, 30.~Nov.~2015).
A \verb!C++! implementation is available at
\url{http://github.com/kzahedi/entropy/}.

Note that we use a compressed notion in Alg.~2--5, in which $x' = x(t+1)$ and $x
= x(t)$.

\begin{algorithm}[h!]
  \begin{algorithmic}[1]
    \REQUIRE{$t=1,2,\ldots,T$}
    \REQUIRE{time series $y=(y(t))$, $\dot{y}=(\dot{y}(t))$,
      $\ddot{y}=(\ddot{y}(t))$, $a=(a(t))$,
      $s=(s(t))$, $t=1,2,\ldots,T$}
    \REQUIRE{Number of bins $B_x$ for time series $x$}
    \STATE{$\discretised{y}(t) = (y(t) - \mathrm{min}(y))/(\mathrm{max}(y) -
      \mathrm{min}(y))) \cdot B_y$}
    \STATE{repeat previous step analogously for $\dot{y}$, $\ddot{y}$, $a$, and
      $s$ to generate discretised time series $\discretised{\dot{y}}$,
      $\discretised{\ddot{y}}$, $\discretised{a}$, and $\discretised{s}$}
    \STATE{$\discretised{w}(t) = \discretised{y}(t) + B_y \cdot
      \discretised{\dot{y}}(t) + B_yB_{\dot{y}} \cdot \discretised{\ddot{y}}(t)$}
    \STATE{The previous step must be applied to sensors and actuators, if they
      result from more than one time series}
    \STATE{$\discretised{w'} = (w^*(2), w^*(3),\ldots, w^*(T))$}
    \STATE{$\discretised{w}  = (w^*(1), w^*(2),\ldots, w^*(T-1))$}
    \STATE{$\discretised{s}  = (s^*(1), s^*(2),\ldots, s^*(T-1))$}
    \STATE{$\discretised{a}  = (a^*(1), a^*(2),\ldots, a^*(T-1))$}
  \end{algorithmic}
  \caption{Discretisation of the data. This part is the same for all measures,
    depending on which time series are required. The $\mathrm{min}$ and
    $\mathrm{max}$ were determined of the data of all hopping models.}%
  \label{alg:discretisation}
\end{algorithm}

\begin{algorithm}[h!]
  \begin{algorithmic}[1]
    \STATE{$p(w',w,a) \leftarrow (0)_{|W|\times|W|\times|A|}$ }\COMMENT{Matrix with
      $|W|\times|W|\times|A|$ entries set to zero}
    \FOR{$t = 1,2,\ldots,T-1$ and $w_{t+1},w_t\in
      \discretised{w},a_t\in\discretised{a}$}
    \STATE{$p(w_{t+1},w_{t},a_{t}) \leftarrow p(w_{t+1},w_{t},a_{t}) + 1$}
    \ENDFOR
    \STATE{$p(w',w,a) \leftarrow p(w',w,a) / (T-1)$}

    \STATE{Estimate $p(w',a)$ from $\discretised{w},\discretised{a}$ or by summing over
      $w$}
    \STATE{$p(w'|w,a)  = \nicefrac{p(w',w,a)}{\sum_{w'}p(w',w,a)}$}
    \STATE{$p(w'|a)    = \nicefrac{p(w',a)}{\sum_{w'}p(w',a)}$}
    \STATE{$\mathrm{MC}_\mathrm{W} = \sum_{w',w,a} p(w',w,a)
     \log_2\frac{p(w'|w,a)}{p(w'|a)}$}
  \end{algorithmic}
  \caption{Algorithm for $\mathrm{MC}_\mathrm{W}$.}\label{alg:mc_w}
\end{algorithm}

\begin{algorithm}[h!]
  \begin{algorithmic}[1]
    \STATE{Perform steps 1--8 from Alg.~\ref{alg:mc_w}}
    \FOR{$t = 1,2,\ldots,T-1$ and $w',w\in
      \discretised{w},a\in\discretised{a}$}
    \STATE{$\mathrm{MC}_\mathrm{W}(t) = \log_2\frac{p(w'|w,a)}{p(w'|a)}$}
    \ENDFOR
  \end{algorithmic}
  \caption{Algorithm for state-dependent
    $\mathrm{MC}_\mathrm{W}(t)$.}\label{alg:sd_mc_w}
\end{algorithm}

\begin{algorithm}[h!]
  \begin{algorithmic}[1]
    \STATE{Estimate $p(w',w)$ from $\discretised{w}$ (see Alg.~\ref{alg:mc_w})}
    \STATE{Estimate $p(a,s)$ from $\discretised{a},\discretised{s}$, (see Alg.~\ref{alg:mc_w})}
    \STATE{$H(W') =  -\sum_{w'}   p(w')   \log_2 p(w')$}
    \STATE{$H(W'|W) = -\sum_{w',w} p(w',w) \log_2
      \nicefrac{p(w',w)}{\sum_{w'} p(w',w)}$}
    \STATE{$H(A) = -\sum_{a}    p(a)    \log_2 p(a)$}
    \STATE{$H(A|S) = -\sum_{a,s}  p(a,s)  \log_2
      \nicefrac{p(a,s)}{\sum_{a} p(a,s)}$}
    \STATE{$\mathrm{MC}_\mathrm{MI} = H(W') - H(W'|W) - H(A) + H(A|S)$}
  \end{algorithmic}
  \caption{Algorithm for $\mathrm{MC}_\mathrm{MI}$.}\label{alg:mc_mi}
\end{algorithm}

\begin{algorithm}[h!]
  \begin{algorithmic}[1]
    \STATE{Perform step 1--2 from Alg.~\ref{alg:mc_mi}}
    \FOR{$t = 1,2,\ldots,T-1$ and $w',w\in
      \discretised{w},a\in\discretised{a},s\in\discretised{s}$}
    \STATE{$\mathrm{MC}_\mathrm{MI}(t) =
      \begin{aligned}[t]
        & \log_2 p(w') - \log_2 p(w'|w)\\
        & + \log_2 p(a) - \log_2 p(a|s)
    \end{aligned}
    $}
    \ENDFOR

  \end{algorithmic}
  \caption{Algorithm for state-dependent
    $\mathrm{MC}_\mathrm{MI}(t)$.}\label{alg:sd_mc_mi}
\end{algorithm}

\section{$I(W';A|W)=0$ for deterministic systems}
\label{app:zero}
In the case where $\alpha(w'|w,a)$,
$\beta(s|w)$, and $\pi(a|s)$ are deterministic, the conditional
entropy $H(W'|W)$ vanishes. It follows that
\begin{align*}
  0 \leq I(W';A|W) & = H(W'|W) - H(W'|W,A)\\
            & \leq H(W'|W) \\
            & = 0.
\end{align*}

\section{Relation between \MCW and \MCMI}
\label{app:relation}
From the following equality
\begin{align*}
  I(W';W,A) & = I(W';W) + I(W';A|W)\\
            & = I(W';A) + I(W';W|A)\\
\end{align*}
we can derive
\begin{align*}
  \underbrace{I(W';W|A)}_{\mathrm{MC}_\mathrm{W}} & = \underbrace{I(W';W) - I(A;S)}_{\mathrm{MC}_\mathrm{MI}} + I(A;S)\\
            & + I(W';A|W) - I(W';A)\\
\mathrm{MC}_\mathrm{W} -
\mathrm{MC}_\mathrm{MI}  & =
I(A;S) + I(W';A|W) - I(W';A)\\
& = H(A) - H(A|S) \\
& \phantom{=} + H(W'|W) - H(W'|W,A)\\
& \phantom{=} - H(A) + H(A|W')\\
& = H(A|W')
\end{align*}
For deterministic systems the conditional entropies 
$H(A|S) = H(W'|W) = H(W'|A,W) = 0$. We show this exemplarily for $H(A|S)$. If
the action $A$ is a function of the sensor state $S$, then $p(a,s) = p(a|s)$ is
either one or zero, because there is exactly one actuator value for every sensor
state. Hence, $H(A|S) = \sum_{a,s} p(a,s) \log p(a|s) = 0$.
The equality $\mathrm{MC}_\mathrm{MI} - \mathrm{MC}_\mathrm{W} = H(A|W')$ is not
hold in Tab.~\ref{tab:results_on_data}, because the discretization introduces
stochasticity, and hence, the conditional entropies are only approximately zero,
i.e., $H(A|S) \approx H(W'|W) \approx H(W'|A,W) \approx 0$

\section*{ACKNOWLEDGMENTS}
This work was partly funded by the DFG Priority Program Autonomous Learning
(DFG-SPP 1527). The DC-motor model is based on a Simulink model provided by
Roger Aarenstrup
(\url{http://in.mathworks.com/matlabcentral/fileexchange/11829-dc-motor-model}).

\onecolumn

\addtolength{\textheight}{-12cm}

\end{document}